\documentstyle[10pt,a4,code,twocolumn]{article}
\scrollmode

\newcommand{\mysection}[1]{\vspace{-1ex}\subsection*{#1}}
\newcommand{\mysubsection}[1]{\vspace{-1ex}\subsubsection*{#1}}
\newcommand{\ndrule}[3]{\mbox{$\frac{\displaystyle #1}{\displaystyle #2}$#3}}

\newcommand{\nrndrule}[2]{\mbox{$\frac{\displaystyle #1}{\displaystyle #2}$}}

\newcommand{\rulegap}[2]{\mbox{\nrndrule{#1}{\nrndrule{\vdots}{#2}}}}

\newcommand{\comment}[1]{}

\def\phi{\varphi}

\newcommand{\und}{\wedge}
\newcommand{\od}{\vee}

\newcommand{\all}{\forall}

\newcommand{\tri}{\triangle}

\newcommand{\mkrp}{{\sc MKRP}}
\newcommand{\omegamkrp}{\mbox{$\Omega$--\mkrp}}
\newcommand{\proverb}{{\em PROVERB\/}}

\def\thebibliography#1{\section*{References}\small\list
  {\arabic{enumi}.}{\settowidth\labelwidth{#1.}\leftmargin\labelwidth
    \advance\leftmargin\labelsep
    \usecounter{enumi}}
    \def\newblock{\hskip .11em plus .33em minus -.07em}
    \sloppy
    \sfcode`\.=1000\relax}

\def\iwg{International Workshop on Natural Language Generation}

\def\seki-repor{SEKI Report}
\def\seki-working-papers{SEKI Working Papers}
\def\uni-sb{Fachbereich Informatik, Universit{\"a}t des Saarlandes}

\def\mitpress{MIT Press}
\def\morgan-kaufmann{Morgan Kaufmann}
\def\ac-press{Academic Press}

\def\proc{Proc.\ }
\def\univ{University}

\title{Planning Argumentative Texts}

\author{Xiaorong Huang\\Fachbereich Informatik, Universit\"at des Saarlandes\\
66041 Saarbr\"ucken, Germany, email: huang@cs.uni-sb.de}

\begin{document}
\date{}

\maketitle \thispagestyle{empty}

\mysection{Abstract} This paper presents \proverb\, a text planner for
argumentative texts. \proverb\'s main feature is that it combines global
hierarchical planning and unplanned organization of text with respect
to local derivation relations in a complementary way.  The former
splits the task of presenting a particular proof into subtasks of
presenting subproofs.  The latter simulates how the next intermediate
conclusion to be presented is chosen under the guidance of the local
focus.

\mysection{1. Introduction}\label{introduction}
This paper presents a text planner for the verbalization of natural
deduction (ND) style proofs \cite{Gentzen35}. Several similar attempts
can be found in previous work. Developed before the era of NL
generation, the system EXPOUND of D.~Chester \cite{Chester76} can be
characterized as an example of {\em direct translation}: Although a
sophisticated linearization is applied on the input ND proofs, the
steps are then translated locally in a template driven way.  ND proofs
were tested as input to an early version of the MUMBLE system of
D.~McDonald \cite{McDonald83}, the main aim however, was to show the
feasibility of the architecture. A more recent attempt can be found in
THINKER \cite{EdPe93}, which implements several interesting but
isolated proof presentation strategies, without giving a comprehensive
underlying model.

Our computational model can therefore be viewed as the first serious attempt
at a comprehensive computational model that produces adequate
argumentative texts from ND style proofs. The main aim is to show how
existing text planning techniques can be adapted for
this particular application. To test its feasibility, this
computational model is implemented in a system called \proverb.

Most current NL text planners assume that language generation is
planned behavior and therefore adopt a hierarchical planning approach
\cite{Hovy88,Moore89,Dale92,Reithinger91}.  Nonetheless there is
psychological evidence that language has an unplanned, spontaneous
aspect as well \cite{Ochs79}.  Based on this observation, researchers
have exploited organizing text with respect to some local relations.
Sibun \cite{Sibun90} implemented a system generating descriptions for
objects with a strong domain structure, such as houses, chips and
families. Once a discourse is started, local structures suggest the
next objects available. Instead of planning globally, short-range
strategies are employed to organize a short segment of text. From a
computational point of view, a hierarchical planner elaborates
recursively on the initial communicative goal until the final
subgoals can be achieved by applying a primitive operator. A text
generator based on the local organization, in contrast, repeatedly chooses a
part of the remaining task and carries it out.

The macroplanner of \proverb\ combines {\em hierarchical
planning} with {\em local organization} in a uniform planning framework.
The hierarchical planning is realized by so-called top-down
presentation operators that split the task of presenting a particular
proof into subtasks of presenting subproofs. While the overall
planning mechanism follows the RST-based planning approach
\cite{Moore89,Reithinger91}, the planning operators more closely resemble the
schemata in {\em schema-based} planning \cite{McKeown85,Paris88}.  Bottom-up
presentation operators are devised to simulate the unplanned aspect,
where the next intermediate conclusion to be presented is chosen
under the guidance of the local focus mechanism in a more
spontaneous way. Since top-down operators embody explicit
communicative norms, they are always given a higher priority. Only
when no top-down presentation operator is applicable, will a bottom-up
presentation operator be chosen.

This distinction between planned and unplanned presentation leads to a
very natural segmentation of the discourse into an {\em attentional
hierarchy\/}, since, following the theory of Grosz and Sidner \cite{GrSi86},
there is a one-to-one correspondence between the intentional
hierarchy and the attentional hierarchy. This attentional hierarchy is
used to make {\em reference choices\/} for inference methods and for
previously presented intermediate conclusions. The inference choices are the
main concern of the microplanner of \proverb (see \cite{Huang94b}).

\mysection{2. Context of Our Research}\label{context}
\begin{figure*}
\begin{center}
\ndrule{\ndrule{u(u_1,1_u,*), u_1\in U}
               {[2]:\ u_1 * 1_u = u_1}
               {Du}
        ,\hspace{1mm}
        \ndrule{u_1\in U, \ndrule{sgr(U,F)}
                                 {U\subset F}
                                 {Dsubgr}}
               {[3]:\ u_1\in F}
               {Ds}
        ,\hspace{1mm}
        \ndrule{U\subset F,\ndrule{u(U,1_u,*)}
                                  {1_u\in U}
                                  {Du}}
               {[4]:\ 1_u\in F}
               {Ds}
        ,\hspace{1mm}
        \ndrule{gr(F,*)}
               {segr(F,*)}
               {Dg}
        ,\hspace{1mm}}
       {[1]:\ Solution(u_1,u_1,1_u,F,*)}
       {Tsol}
\end{center}
\caption{An Example Input Proof\label{ex-input}}
\end{figure*}

The text planner discussed in this paper is the macroplanner of
\proverb, which translates machine-found proofs in several steps into
natural language. \proverb\ adopts a {\em reconstructive} approach:
Once a proof in a machine oriented formalism is generated in the proof
development environment \omegamkrp, a new proof that more resembles
those found in mathematical textbooks is reconstructed
\cite{Huang94a}. The reconstructed proof is a {\em proof tree\/},
where {\em proof nodes\/} are derived from their children by applying
an inference method (also called a justification). Most of the steps
are justified by the application of a definition or a theorem, the
rest are justified by inference rules of the natural deduction (ND)
calculus, such as the ``Case'' rule.  Figure \ref{ex-input} is an
example of a segment of a possible input proof, where some nodes are
labeled for convenience.

\noindent The justifications ``Du'', ``Dsubgr'', ``Ds'', ``Dg'', and
``Tsol'' stand for the definitions of unit element, of subgroup,
of subset, of group, and the theorem about solution, respectively.

The input proof tree is also augmented with an ordered list of nodes,
being roots of subproofs planned in this order. The proof in Figure
\ref{ex-input} is associated with the list: ([2], [3], [4], [1]).

\mysection{3. The Framework of the Macroplanner}\label{framework}
The macroplanner of \proverb\ elaborates on communicative goals,
selects and orders pieces of information to fulfill these goals.  The
output is an ordered sequence of {\em proof communicative act
  intentions\/} (PCAs). PCAs can be viewed as {\em speech acts\/} in our domain
of application.

\mysubsection{Planning Framework}
\proverb\ combines the two above mentioned presentation modes by
encoding communication knowledge for both top-down planning and
bottom-up presentation in form of operators in a uniform planning
framework. Since top-down presentation operators embody explicit
communicative norms, they are given a higher priority. A bottom-up
presentation is chosen only when no top-down presentation operator
applies. The overall planning framework is realized by the function
{\tt Present\/}. Taking as input a subproof, {\tt Present} repeatedly
executes a basic planning cycle until the input subproof is conveyed.
Each cycle carries out one presentation operator, where {\tt Present}
always tries first to choose and apply a top-down operator. If
impossible, a bottom-up operator will be chosen.  The function {\tt
  Present} is first called with the entire proof as the presentation
task. The execution of a top-down presentation operator may generate
subtasks by calling it recursively.  The discourse produced by each
call to {\tt Present} forms an attentional unit (compare the
subsection below).

\mysubsection{The Discourse Model and the Attentional Hierarchy}
The discourse carried out so far is recorded in a {\em discourse
  model\/}.  Rather than recording the semantic objects and their
properties, our discourse
model consists basically of the part of the input proof tree which has
already been conveyed.  The discourse model is also segmented into an
{\em attentional hierarchy}, where subproofs posted by a top-down presentation
operators as subtasks constitute attentional units.
The following are some notions useful for the formulation of the
presentation operators:

\begin{itemize}
\item {\em Task\/} is the subproof in the input proof whose
  presentation is the current task.
\item {\em Local focus\/} is the intermediate conclusion last
  presented, while the semantic objects involved in the local focus are
  called the {\em focal centers}.
\end{itemize}

\vspace{-3mm}
\mysubsection{Proof Communicative Acts}\label{pcas}
PCAs are the primitive actions planned during the macroplanning to
achieve communicative goals. Like speech acts, PCAs can be defined in
terms of the communicative goals they fulfill as well as their
possible verbalizations.  Based on an analysis of
proofs in mathematical textbooks, each PCA has as goal a combination
of the following subgoals:

\begin{enumerate}
\item Conveying a step of the derivation. The simplest PCA is the operator {\tt
Derive}. Instantiated as below:

\vspace{-2mm}
\begin{code}
(Derive Reasons: {\dcd ($a\in S_1$, $S_1\subseteq S_2$)}
        Intermediate-Results: nil
        Derived-Formula: {\dcd $a\in S_2$}
        Method: def-subset)
\end{code}

\vspace{-5mm} depending on the reference choices, a possible
verbalization is given as following:

\vspace{-1mm}\begin{quote}
  ``Because $a$ is an element of $S_1$ and $S_1$ is a subset of
  $S_2$, according to the definition of subset, $a$ is an element of
  $S_2$.''
\vspace{-1mm}\end{quote}

\item Updates of the global attentional structure.
  These PCAs sometimes also convey a partial plan for the further
  presentation. Effects of this group of PCAs include: creating new
  attentional units, setting up partially premises and the goal of a new
  unit, closing the current unit, or reallocating the attention of the
  reader from one attentional unit to another. The PCA

\begin{code}
(Begin-Cases Goal: {\dcd $\it Formula$}
             Assumptions: {\dcd ($A$ $B$)})
\end{code}

\vspace{-5mm}
\noindent creates two attentional units with $A$ and $B$ as the
assumptions, and {\it Formula\/} as the goal by producing the
verbalization:

\vspace{-1mm}\begin{quote}
``To prove {\it Formula\/}, let us consider the two
cases by assuming $A$ and $B$.''
\vspace{-1mm}\end{quote}
\end{enumerate}

Thirteen PCAs are currently employed in \proverb. See
\cite{Huang94} for more details.

\mysubsection{Structure of the Planning Operators}
Although top-down and bottom-up presentation activities are of a
conceptually different nature, the corresponding communication
knowledge is uniformly encoded as {\em presentation operators} in a
planning framework, similar to the plan operators in other generation
systems \cite{Hovy88,Moore89,Dale92,Reithinger91}. In general,
presentation operators map an original presentation task into a
sequence of subtasks and finally into a sequence of PCAs.
\comment{They embody either general communicative norms or
  communicative norms specific for argumentative text, especially for
  mathematical proofs. As such, presentation operators ensure the
  coherence of the text produced.} All of them have the following four
slots:
\begin{itemize}
\item {\em Proof}: a proof schema, which characterizes the syntactical
  structure of a proof segment for which this operator is designed.
  It plays the role of the goal slot in the traditional planning
  framework.
\item {\em Applicability Condition\/}: a predicate.
\item {\em Acts}: a procedure which essentially carries out a sequence
  of presentation acts. They are either primitive PCAs, or are
  recursive calls to the procedure {\tt Present} for subproofs.
\item {\em Features}: a list of features which helps to select the best
  of a set of applicable operators.
\end{itemize}

\mysection{4. Top-Down Planning}\label{top-down}
This section elaborates on the communicative norms concerning how a
proof to be presented can be split into {\em subproofs\/}, as well as
how the hierarchically-structured subproofs can be mapped onto some
{\em linear} order for presentation.  In contrast with operators employed
in RST-based planners that split goals according to the rhetorical
structures, our operators encode standard schemata for presenting
proofs, which contain subgoals. The top-down
presentation operators are roughly divided into two categories:

\begin{itemize}
\item schemata-based operators encoding complex schemata for the
  presentation of proofs of a specific pattern (twelve of them are currently
integrated in \proverb),
\item general operators embodying general presentation norms,
  concerning splitting proofs and ordering subgoals.
\end{itemize}

\vspace{-3mm}
\begin{figure}[h]
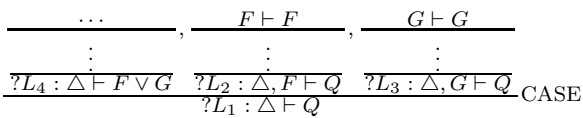

\begin{center}\footnotesize
\ndrule{\rulegap{\cdots}{?L_4:\tri\vdash F\vee G},
  \rulegap{F\vdash F}{?L_2: \tri, F\vdash Q},
  \rulegap{G\vdash G}{?L_3: \tri, G\vdash Q}}{?L_1:\tri\vdash
  Q}{CASE}
\end{center}
\vspace{-2mm}
\caption{A Schema Involving Cases\label{part2-proof-case}}
\end{figure}

Let us first look at an operator devised for proof segments containing
cases. The corresponding schema of such a proof tree is shown in
Figure \ref{part2-proof-case}. Under two circumstances a writer may
recognize that he is confronted with a proof segment containing cases.
First, when the subproof that has the structure of Figure
\ref{part2-proof-case} is the current presentation task, tested by
(task $?L_1$)\footnote{Labels stand for the corresponding nodes}.
Second, when the disjunction $F\vee G$ has just been presented in the
bottom-up mode, tested by (local-focus $?L_4$).  Under both
circumstances, a communication norm motivates the writer to first
present the part leading to $F\vee G$ (in the second case this subgoal
has already been achieved), and then to proceed with the two cases. It
enforces also that certain PCAs be used to mediate between
parts of proofs. This procedure is exactly captured by the
presentation operator below.

\noindent
{\bf Case-Implicit}
\begin{itemize}
\item Proof: as given in Figure \ref{part2-proof-case}
\item Applicability Condition:  ((task $?L_1$) $\vee$\\ (local-focus $?L_4$))
$\wedge$ (not-conveyed ($?L_2$ $?L_3$))
\item Acts:
\begin{enumerate}
\item if $?L_4$ has not been conveyed,
then present $?L_4$ (subgoal 1)
\item a PCA with the verbalization: ``First, let us consider the first case by
  assuming $F$.''
\item present $?L_2$ (subgoal 2)
\item a PCA with the verbalization: ``Next, we consider the second
  case by assuming $G$.''
\item present $?L_3$ (subgoal 3)
\item mark $?L_1$ as conveyed
\end{enumerate}
\item features: (top-down compulsory implicit)
\end{itemize}

The feature values can be divided into two groups: those
characterizing the style of the text this operator produces, and
those concerning other planning aspects. ``Implicit'' is a stylistic
feature value, indicating that the splitting of the proof into the
three subgoals is not made explicit. In its explicit
dual {\bf Case-Explicit} a PCA is added to the beginning of the Acts slot,
which produces the verbalization:

\vspace{-1mm}\begin{quote}
  ``To prove $Q$, let us first prove $F\od G$, and consider the two
  cases separately.''
\vspace{-1mm}\end{quote}

The feature value ``compulsory'' indicates that if the applicability
condition is satisfied, and the style of the operator conforms to the
global style the text planner is committed to, this operator should be
chosen. Two weaker values also reflect the specificity of plan
operators: ``specific'' and ``general''.

General presentation operators perform a simple task according to some
general text organization principles. They either

\begin{itemize}
\item enforce a linearization on subproofs to be presented, or
\item split the task of the presentation of a proof with ordered
  subproofs into subtasks.
\end{itemize}

The first ordering operator operationalizes a general ordering
strategy called {\em minimal load principle}.  This principle
predicates that a writer usually presents shorter branches before
longer ones. The argument of Levelt is rather simple: When one branch
is chosen to be described first, the writer has to have the {\em
  choice node\/} flagged in his memory for return. If he follows the
shorter branch first, the duration of the load will be shorter. The
concrete operator is omitted.

Note that, the subproofs being ordered are subproofs conceptually planned
while the corresponding proof is constructed. There are two other
ordering operators based on general ordering principles: the {\em
  local focus\/} principle and the {\em proof time order\/} principle
\cite{Huang94}.

The invocation of an ordering operator is always followed by the
invocation of a splitting operator, which actually posts subgoals by calling
the function {\tt Present} with the ordered goals subsequently.

\mysection{5. Bottom-up Presentation}\label{bottom-up}
The {\em bottom-up presentation\/} process simulates the unplanned
part of proof presentation. Instead of splitting presentation goals
into subgoals according to standard schemata, it follows the local
derivation relation to find a next proof node or subproof to be
presented. In this sense, it is similar to the local organization
techniques used in \cite{Sibun90}. When no top-down presentation
operator applies, \proverb\ chooses a bottom-up operator.

\mysubsection{The Local Focus}\label{local-focus}
The node to be presented next is suggested by the mechanism of {\em
  local focus\/}. Although logically any proof node having the local
focus as a child could be chosen for the next step, usually the one
with the greatest semantic overlapping with the {\em focal centers\/}
is preferred. As mentioned above, focal centers are semantic objects
mentioned in the proof node which is the local focus.  This is based
on the observation that if one has proved a property about some
semantic objects, one tends to continue to talk about these
particular objects before turning to new objects. Let us
examine the situation when the proof below is awaiting presentation.

\begin{center}
\nrndrule{\nrndrule{[1]:\ P(a,b)}{[2]:\ Q(a,b)}
          ,\hspace{2cm}
          \nrndrule{[1]:\ P(a,b),\ [3]:\ S(c)}{[4]:\ R(b,c)}}
         {[5]:\ Q(a,b)\und R(b,c)}
\end{center}

Assume that node [1] is the local focus, the set $\{a,b\}$ are the
focal centers, [3] is a previously presented node and node [5] is the
current task.  [2] is chosen as the next node to be presented, since
it does not (re)introduce any new semantic object and its overlap
with the focal centers ($\{a,b\}$) is larger than those of [4] ($\{b\}$).

\mysubsection{The Bottom-Up Presentation Operators}
Under different circumstances the derivation of the next-node is also
presented in different ways. The corresponding presentation knowledge
is encoded as bottom-up presentation operators. The one most frequently
used  presents one
step of derivation:

\pagebreak[3]
\vspace{0.3cm}\noindent
{\bf Derive-Bottom-Up}
\nopagebreak[4]
\begin{itemize}
\item Proof: \ndrule{{\it ?Node}_1, \ldots, {\it ?Node}_n}{{\it
?Node}_{n+1}}{{\it ?M}}
\item Applicability Condition:
      $?{\it Node}_{n+1}$ is suggested by the focus mechanism as the next node,
and  $?{\it Node}_1$, $\ldots$, $?{\it Node}_n$ are conveyed.
\item Acts: a PCA that conveys the fact that $?{\it Node}_{n+1}$ is derived
from the premises $?{\it Node}_1$, $\ldots$, $?{\it Node}_n$ by applying ${\it
?M}$.
\item Features: (bottom-up general explicit detailed)
\end{itemize}

If the conclusion $?{\it Node}_{n+1}$, the premises and the method
${\it ?M}$ are instantiated to $a\in S_1$, ($a\in S_2$, $S_1\in S_2$), {\it
  def-subset} respectively, the following verbalization can be
produced:

\vspace{-1mm}\begin{quote}
  ``Since $a$ is an element of $S_1$, and $S_1$ is a subset of $S_2$,
  $a$ is an element of $S_2$ according to the definition of subset.''
\vspace{-1mm}\end{quote}

A {\em trivial\/} subproof may be  presented as a single derivation by omitting
the
intermediate nodes. This {\em next subproof\/} is also suggested by the local
focus. This is simulated by a bottom-up operator called {\bf
Simplify-Bottom-Up}. Currently seven bottom-up operators are integrated
in \proverb.

\comment{
In \proverb\ it is required that the {\tt next-node} is in this
subproof and this subproof is part of the current task. The
presentation activities discussed above is captured in the
presentation operator below.

\vspace{0.5cm}\noindent
{\bf Simplify-Bottom-Up}
\begin{itemize}
\item Proof: \nrndrule{?{\it Node}_1, \ldots, ?{\it Node}_n}
                      {\nrndrule{\vdots}
                                {?{\it Node}_{n+1}}}
\item Applicability Condition:
(in-subproof {\it ?Node}$_{n+1}$ task)
      $\und $ (in-subproof  next-node ${\it Node}_{n+1}$)
      $\und \all_{i=1\ to\ n}$ (conveyed $?{\it Node}_i$)
      $\und$ (simple-proof $?{\it Node}_{n+1})$
\item Acts:
\begin{code}
(Derive Derived-Formula: {\dcd$?Node\sb{n+1}$},
        Reasons: {\dcd $?{\it Node}_1,\ldots,?{\it Node}_n$)}
\end{code}
\item Features (bottom-up general explicit abstract)
\end{itemize}

\vspace{3mm} The simplicity of a subproof is currently judged by a
heuristic function which takes both the number of nodes and the type
of the derivations in the subproof into account. Unlike the operator
{\bf Derive-Bottom-Up}, no inference method will be mentioned in the
PCA {\tt Derive}.
}

\mysection{6. Verbalization of PCAs}\label{example}
Macroplanning produces a sequence of PCAs. Our microplanner is
restricted to the treatment of the {\em reference choices\/} for the
inference methods and for the previously presented intermediate
conclusions. While the former depends on static salience relating to
the domain knowledge, the latter is similar to
subsequent references, and is therefore sensitive to the context, in
particular to its segmentation into attentional hierarchy. Due to
space restrictions, we only show the following piece of a {\em preverbal
  message\/} as an example, being a PCA enriched with reference
choices for reasons and method by the microplanner \cite{Huang94b,Huang94}.

\begin{code}
(Derive Reasons: (((ELE a U) explicit)
                  ((SUBSET U F) omit))
        Conclusion: (ELE a F)
        Method: (Def-Subset omit))
\end{code}

\vspace{-5mm}
Our surface generator TAG-GEN \cite{Kilger93} produces the utterance:

\vspace{-1mm}\begin{quote}
``Since $a$ is an element of $U$, $a$ is an element
of $F$.''
\vspace{-1mm}\end{quote}

Notice, only the reason labeled as ``explicit'' is verbalized.

Finally, to demonstrate the type of proofs currently generated by
\proverb, below is the complete output for a proof constructed by
\omegamkrp:

\vspace{2mm}
\noindent {\bf Theorem}: Let $F$ be a group and $U$ a subgroup of $F$,
if $1$ and $1_U$ are unit elements of $F$ and $U$ respectively, then
$1=1_U$.

\pagebreak[3]
\noindent {\bf Proof:}
\nopagebreak[4]

Let $F$ be a group, $U$ be a subgroup of $F$, $1$ be a unit element of $F$ and
$1_U$ be a unit element of $U$.
According to the definition of unit element, $1_U \in U$.
Therefore there is an $X$, $X \in U$.
Now suppose that $u_1$ is such an $X$.
According to the definition of unit element, $u_1*1_U = u_1$.
Since $U$ is a subgroup of $F$, $U \subset F$.
Therefore $1_U \in F$.
Similarly $u_1 \in F$, since $u_1 \in U$.
Since $F$ is a group, $F$ is a semigroup.
Because $u_1*1_U = u_1$, $1_U$ is a solution of the equation $u_1*X=u_1$.
Since $1$ is a unit element of $F$, $u_1*1 = u_1$.
Since $1$ is a unit element of $F$, $1 \in F$.
Because $u_1 \in F$, $1$ is a solution of the equation $u_1*X=u_1$.
Since $F$ is a group, $1_U = 1$ by the uniqueness of solution.
This conclusion is independent of the choice of the element $u_1$.

\mysection{7. Conclusion and Future Work}\label{conclusion}
This paper puts forward an architecture that combines several
established NL generation techniques adapted for a particular
application, namely the presentation of ND style proofs. We hope that
this architecture is also of general interest beyond this particular
application.

The most important feature of this model is that hierarchical
planning and unplanned spontaneous presentation are integrated in a
uniform framework. Top-down hierarchical planning views language
generation as planned behavior.  Based on explicit communicative
knowledge encoded as schemata, hierarchical planning splits a
presentation task into subtasks. Although our overall presentation
mechanism has much in common with that of RST-based text planners, the
top-down planning operators contain mostly complex presentation
schemata, like those in schema-based planning. Since schemata-based
planning covers only proofs of some particular structure, it is
complemented by a mechanism called bottom-up presentation. Bottom-up
presentation aims at simulating the unplanned part of proof
presentation, where a proof node or a subproof awaiting presentation
is chosen as the next to be presented via the local derivation
relations. Since more than one such node is often available, the local
focus mechanism is employed to single out the candidate having
the strongest semantic links with the focal centers. The distinction
between planned and unplanned behavior enables a very natural
segmentation of the discourse into an attentional hierarchy. This
provide an appropriate basis for a discourse theory which handles
reference choices \cite{Huang94b}.

Compared with proofs found in mathematical textbooks, the output of
\proverb\ is still too tedious and inflexible. The tediousness is
largely ascribed to the lack of {\em plan level\/} knowledge of the
input proofs, which distinguishes crucial steps from unimportant
details.  Therefore, sophisticated plan recognition techniques are
necessary.  The inflexibility of text currently produced is partly
inherited from the schemata-based approach, for which a fine-grained
planning in terms of single PCAs might be a remedy. It is also partly
due to the fixed lexicon choice, which we are currently
reimplementing.

\bibliographystyle{alpha}

\begin{thebibliography}{HKK{\etalchar{+}}ng}

\bibitem[Che76]{Chester76}
D. Chester.
\newblock The translation of formal proofs into {E}nglish.
\newblock {\em Artificial Intelligence}, 1976.

\bibitem[Dal92]{Dale92}
R.~Dale.
\newblock {\em Generating Referring Expressions}.
\newblock \mitpress, 1992.

\bibitem[EP93]{EdPe93}
A. Edgar and F.~J.~Pelletier.
\newblock Natural language explanation of natural deduction proofs.
\newblock In {\em \proc of the first Conf. of the Pacific Assoc. for
  Comp. Linguistics}, 1993.

\bibitem[Gen35]{Gentzen35}
G. Gentzen.
\newblock Untersuchungen \"uber das logische Schlie{\ss}en I.
\newblock {\em Math. Zeitschrift}, 1935.

\bibitem[GS86]{GrSi86}
B.~J. Grosz and C.~L. Sidner.
\newblock Attention, intentions, and the structure of discourse.
\newblock {\em Computational Linguistics}, 1986.

\bibitem[Hov88]{Hovy88}
E.~H. Hovy.
\newblock {\em Generating Natural Language under Progmatic Constrints}.
\newblock Lawrence Erlbaum Associates, Hillsdale, 1988.

\bibitem[Hua94a]{Huang94a}
X.~Huang.
\newblock Reconstructing proofs at the assertion level.
\newblock In {\em \proc of 12th CADE}, 1994, forthcoming.

\bibitem[Hua94b]{Huang94b}
X.~Huang.
\newblock Planning Reference Choices for Argumentative Texts.
\newblock In {\em \proc of the 7th \iwg}, 1994, forthcoming.

\bibitem[Hua94b]{Huang94}
X.~Huang.
\newblock {\em A Reconstructive Approach to Human Oriented Proof Presentation}.
\newblock PhD thesis, Universit\"at des Saarlandes, Germany, 1994, forthcoming.

\bibitem[Kil94]{Kilger93}
A. Kilger.
\newblock Using {UTAG}s for incremental and parallel generation.
\newblock {\em Computational Intelligence}, forthcoming, 1994.

\bibitem[McD83]{McDonald83}
D.~D.~McDonald.
\newblock Natural language generation as a computational problem.
\newblock In {\em Brady/Berwick: Computational Models of Discourse}. \mitpress,
  1983.

\bibitem[McK85]{McKeown85}
K.~R. McKeown.
\newblock {\em Text Generation}.
\newblock Cambridge \univ\ Press, 1985.

\bibitem[Moo89]{Moore89}
J.~D.~Moore.
\newblock {\em A Reactive Approach to Explanation in Expert and Advice-Giving
  Systems}.
\newblock PhD thesis, Univ. of California, 1989.

\bibitem[Och79]{Ochs79}
E.~Ochs.
\newblock Planned and unplanned discourse.
\newblock {\em Syntax and Semantics}, 1979.

\bibitem[Par88]{Paris88}
C. Paris.
\newblock Tailoring object descriptions to a user's level of expertise.
\newblock {\em Computational Linguistics}, 1988.

\bibitem[Rei91]{Reithinger91}
N.~Reithinger.
\newblock {\em Eine parallele Architektur zur inkrementeller Dialogbeitr\"age}.
\newblock PhD thesis, Universit\"at des Saarlandes, 1991.

\bibitem[Sib90]{Sibun90}
P. Sibun.
\newblock The local organization of text.
\newblock In K.R. McKeown etal, editors, {\em \proc\
  of the 5th \iwg}, 1990.

\end{thebibliography}
\newcommand{\etalchar}[1]{$^{#1}$}

\end{document}